\documentclass[journal]{IEEEtran}
\ifCLASSINFOpdf
\else
\fi
\usepackage{amsmath}
\usepackage[pdftex]{graphicx}
\usepackage{times}
\usepackage{caption2}
\usepackage{amsmath}
\usepackage{amsfonts}
\usepackage{amssymb}
\usepackage[mathscr]{euscript}
\usepackage{bm}
\usepackage{subfigure}
\usepackage{algorithm}
\usepackage{algorithmic}
\usepackage{multirow}
\usepackage{wrapfig}
\usepackage{booktabs}
\usepackage{threeparttable}
\usepackage[english]{babel}
\usepackage{microtype}
\usepackage{epsfig}
\usepackage{graphicx}

\DeclareMathOperator*{\argmin}{argmin}

\newcommand{\paraheading}[1]{\textbf{{#1}.~}}

\usepackage[bookmarks,backref=true,linkcolor=black]{hyperref}
\hypersetup{
  pdfauthor = {},
  pdftitle = {},
  pdfsubject = {},
  pdfkeywords = {},
  colorlinks=true,
  linkcolor= black,
  citecolor= black,
  pageanchor=true,
  urlcolor = black,
  plainpages = false,
  linktocpage
}
\usepackage{color}

\definecolor{turquoise}{cmyk}{0.65,0,0.1,0.1}
\definecolor{purple}{rgb}{0.65,0,0.65}
\definecolor{dark_green}{rgb}{0, 0.5, 0}
\definecolor{orange}{rgb}{0.8, 0.2, 0.2}

\newcommand{\videoinfo}{\footnote{\url{https://www.youtube.com/watch?v=eJOEMlWKybI}}}

\begin{document}
%
% paper title
% Titles are generally capitalized except for words such as a, an, and, as,
% at, but, by, for, in, nor, of, on, or, the, to and up, which are usually
% not capitalized unless they are the first or last word of the title.
% Linebreaks \\ can be used within to get better formatting as desired.
% Do not put math or special symbols in the title.
\title{Deep Face Feature for Face Alignment}

% author names and affiliations
% transmag papers use the long conference author name format.

\author{Boyi Jiang, Juyong~Zhang, Bailin~Deng, Yudong~Guo and~Ligang~Liu
\thanks{Boyi~Jiang, Juyong~Zhang(Corresponding author), Yudong Guo, Ligang Liu are with School of Mathematical Sciences, University of Science and Technology of China. E-mail: \url{jby1993@mail.ustc.edu.cn}, \url{juyong@ustc.edu.cn}, \url{gyd2011@mail.ustc.edu.cn}, \url{lgliu@ustc.edu.cn}.} %
\thanks{Bailin~Deng is with School of Computer Science and Informatics, Cardiff University. E-mail: \url{DengB3@cardiff.ac.uk}.} %
}

% The paper headers
%\markboth{~}{~}
\markboth{~}
{Jiang \MakeLowercase{\textit{et al.}}: Deep Face Feature for Face Alignment}
% The only time the second header will appear is for the odd numbered pages
% after the title page when using the twoside option.
%
% *** Note that you probably will NOT want to include the author's ***
% *** name in the headers of peer review papers.                   ***
% You can use \ifCLASSOPTIONpeerreview for conditional compilation here if
% you desire.

% If you want to put a publisher's ID mark on the page you can do it like
% this:
%\IEEEpubid{0000--0000/00\$00.00~\copyright~2014 IEEE}
% Remember, if you use this you must call \IEEEpubidadjcol in the second
% column for its text to clear the IEEEpubid mark.

% use for special paper notices
%\IEEEspecialpapernotice{(Invited Paper)}

% for Transactions on Magnetics papers, we must declare the abstract and
% index terms PRIOR to the title within the \IEEEtitleabstractindextext
% IEEEtran command as these need to go into the title area created by
\maketitle
% As a general rule, do not put math, special symbols or citations
% in the abstract or keywords.
%\IEEEtitleabstractindextext{%
\begin{abstract}
In this paper, we present a deep learning based image feature extraction method designed specifically for face images. To train the feature extraction model, we construct a large scale photo-realistic face image dataset with ground-truth correspondence between multi-view face images, which are synthesized from real photographs via an inverse rendering procedure. The deep face feature (DFF) is trained using correspondence between face images rendered from different views. Using the trained DFF model, we can extract a feature vector for each pixel of a face image, which distinguishes different facial regions and is shown to be more effective than general-purpose feature descriptors for face-related tasks such as matching and alignment. Based on the DFF, we develop a robust face alignment method, which iteratively updates landmarks, pose and 3D shape. Extensive experiments demonstrate that our method can achieve state-of-the-art results for face alignment under highly unconstrained face images.
\end{abstract}

% Note that keywords are not normally used for peerreview papers.

% make the title area
\maketitle

% To allow for easy dual compilation without having to reenter the
% abstract/keywords data, the \IEEEtitleabstractindextext text will
% not be used in maketitle, but will appear (i.e., to be "transported")
% here as \IEEEdisplaynontitleabstractindextext when the compsoc
% or transmag modes are not selected <OR> if conference mode is selected
% - because all conference papers position the abstract like regular
% papers do.
\IEEEdisplaynontitleabstractindextext
% \IEEEdisplaynontitleabstractindextext has no effect when using
% compsoc or transmag under a non-conference mode.

% For peer review papers, you can put extra information on the cover
% page as needed:
% \ifCLASSOPTIONpeerreview
% \begin{center} \bfseries EDICS Category: 3-BBND \end{center}
% \fi
%
% For peerreview papers, this IEEEtran command inserts a page break and
% creates the second title. It will be ignored for other modes.
\IEEEpeerreviewmaketitle

\section{Introduction}

Face alignment from images has been an active research topic since 1990s. Face alignment plays an important role in many applications such as 3D face reconstruction~\cite{liu2016joint} and face recognition~\cite{wagner2012toward}, because it is often used as a pre-processing step. Recently, face alignment has gained significant progress in both theory and practice. Although AAM-based approaches~\cite{tzimiropoulos2013optimization,saragih2007nonlinear,
cootes2001active,matthews2004active} and regression-based approaches~\cite{cao2014face,ren2014face,xiong2013supervised,zhu2015face} work well for face images with small poses, they usually cannot handle profile face images as they do not consider the visibility of landmarks.

In recent years, several methods introduce the 3D Morphable Model (3DMM)~\cite{blanz1999morphable} for face alignment and achieve better results~\cite{jourabloo2015pose,liu2016joint}. Using a 3D face model to compute the visibility and position of 2D landmarks, these methods can handle challenging cases with large pose variation. However, the reconstruction accuracy of such methods is often insufficient. Existing approaches usually use general-purpose features such as SIFT~\cite{lowe2004distinctive} to determine the parameters of a 3DMM by cascaded regression. On the one hand, 3DMM shape and expression parameters are highly non-linear with image texture information, making the mapping difficult to estimate. On the other hand, SIFT type descriptors are designed based on color information of local patches, which works well for general objects but do not make use of specific priors for face images. Therefore, it is possible to achieve better alignment performance by designing a feature descriptor tailored for face images .

Recently, convolution neural networks (CNN) have been successfully applied to many related tasks with state-of-the-art results. In~\cite{zhu2016face,jourabloo2016large,yu2017learning}, CNN are applied to improve the regression accuracy. In this paper, we propose to use CNN on face images to train a feature extraction model. For each pixel of a face image, we can use this model to extract a high-dimensional feature, which can accurately indicate the same anatomical facial point across different face images under unconstrained conditions, with better performance on many face related tasks than classical features. To train the model, we need a large number of unconstrained face images with registered ground-truth 3DMM faces. However, it is not easy to obtain real face images with ground-truth 3D shapes, especially for profile view face images. To solve this problem, we synthesize a large-scale face image dataset with different poses and expressions together with ground-truth 3D shapes. With the well constructed training set, a novel feature learning method is proposed to extract the feature vector of each pixel such that the feature vector is distinguishable for each face part and smooth over the whole face area. Based on the trained face feature, we design a simple cascaded regression method to perform face alignment in the wild. The experimental results demonstrate that the trained deep face feature has good performance on feature matching, and the face alignment method outperforms existing methods especially on face images with large pose. In summary, the main contributions of this work include:
\begin{itemize}
\item We propose a CNN-based face image feature extraction method using a well-constructed training dataset and a novel loss function; the trained feature outperforms general-purpose feature descriptors such as SIFT.
\item With the newly designed face feature extractor, we propose a simple yet effective cascaded regression-based approach for face alignment in the wild, and the experimental results demonstrate its better performances over existing methods. 
\end{itemize}
%------------------------------------------------------------------------

\begin{figure*}
	\centering
	%\fbox{\rule{0pt}{2in} \rule{.9\linewidth}{0pt}}
	\includegraphics[width=0.9\textwidth]{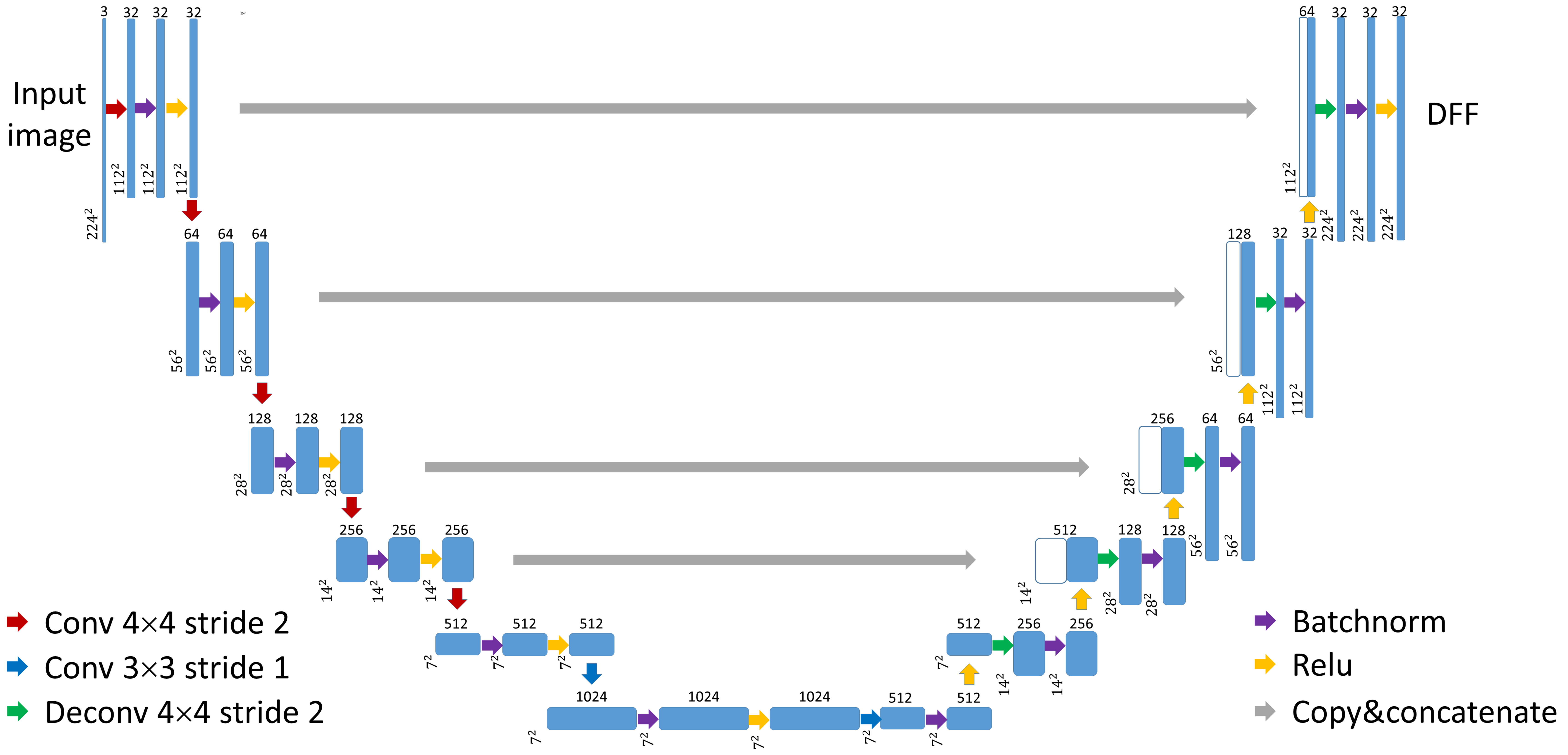}
	\caption{The neural network architecture of DFF extractor. Each blue box represents a multi-channel feature map. The number of channels is on top of the box. The row and column numbers are provided at the lower left edge of the box. White boxes represent copied feature maps.}
	\label{fig:DFF_extractor}
\end{figure*}

%------------------------------------------------------------------------
\section{Related Works}

\paraheading{Classical Face Alignment} Classical face alignment methods, including Active Shape Model (ASM)~\cite{tecootes1994active,cristinacce2007boosted} and Active Appearance Model (AAM)~\cite{cootes2001active,saragih2007nonlinear,tzimiropoulos2013optimization,matthews2004active}, simulate the image generation process and perform face alignment by minimizing the difference between the model appearance and the input image. These methods can achieve accurate reconstruction results, but require a large number of face models with detailed and accurate point-wise correspondence, as well as high computation cost of parameter fitting. Constrained Local Model (CLM)~\cite{asthana2013robust,cristinacce2008automatic} employs discriminative local texture models to regularize the landmark locations. The CLM algorithm is more robust than the AAM method, which updates the model parameters by minimizing the image reconstruction error. Recently, regression based methods~\cite{burgos2013robust, cao2014face} have been proposed to directly estimate landmark locations from the discriminative features around landmarks. Most regression based algorithms do not consider the visibility of facial landmarks under different view angles. As a result, their performance can degrade substantially for input face images with large poses.

\paraheading{Face Alignment via Deep Learning} In recent years, deep learning based methods have been successfylly applied to face alignment and achieved remarkable results. The methods of \cite{sun2013deep,zhang2014coarse,zhou2013extensive} use multi-stage CNN to regress sparse face image landmark locations. To boost the performance of face alignment, Zhang et al.~\cite{zhang2014facial} combine face detection, face alignment, and other tasks into the training of CNN. Alignment of faces with large pose variation is a very challenging problem, because each face might have a different set of visible landmarks. Early works for large-pose face alignment rely on MVS based methods~\cite{yu2013pose,zhu2012face}, which use different landmark templates for different views and incur high computation costs. Recently, 3D model based techniques~\cite{jeni2015dense,jourabloo2015pose,jourabloo2016large,zhu2016face} have been proposed to address the problem of alignment accuracy for challenging inputs, e.g. those with non-frontal face poses, low image quality, and occlusion, etc. These techniques utilize a 3D morphable model~\cite{blanz1999morphable,blanz2003face} (3DMM) to handle self-occluded landmarks and large-pose landmark detection. Jourabloo and Liu~\cite{jourabloo2016large} integrate 2D landmark estimation into the 3D face model fitting process, use cascaded CNN to replace simple regressor, and are able to detect 34 landmarks under all view angles. Zhu et al.~\cite{zhu2016face} reduce the CNN regression complexity by estimating a Projected Normalized Coordinate Code map, which can detect 68 anatomical landmarks with visibility judgement. Yu et al.~\cite{yu2017learning} predict dense facial correspondences by training a encoder-decoder network with synthesized face images, and produce robust face image alignment results.

%%$\mathbf{Large Pose Face Alignment}$:
%The most common approaches in large-pose face alignment are multi-view based method \cite{yu2013pose,zhu2012face}. These methods use different landmark templates for different views, which cause high computation cost because of best template testing and do not detect invisible self-occluded landmarks. Recently, 3D model based techniques \cite{jeni2015dense,jourabloo2015pose,jourabloo2016large,zhu2016face}, to address the problems of alignment
%accuracy, e.g. non-frontal face pose, low image quality, occlusion,
%etc. These techniques utilize a 3D morphable model to handle self-occluded landmarks and large-pose landmarks detection.
%These techniques achieved the state-of-the-art performance with larger poses.

\paraheading{3D Face Reconstruction} The 3DMM establishes statistical linear parametric models for both texture and shapes of human faces, with a 3D face represented by a set of coefficients for its shape and texture basis. To recover the face from a 2D image, 3DMM-based methods~\cite{blanz1999morphable,romdhani2005estimating,blanz2003face} estimate the shape and texture coefficients by maximizing the similarity between the input 2D face image and the projected 3D face. However, such methods are not robust enough to handle facial landmarks under large pose variation. Multi-view stereo (MVS)~\cite{wu2011multicore,wu2013towards} is a classical reconstruction method that requires dense correspondence between neighboring images to achieve satisfactory results. When such methods are applied on multi-view face images, the reconstructed face point cloud might contain holes due to insufficient detected matched points.

%------------------------------------------------------------------------
\section{Our Method}
To train the DFF, we first build a large-scale training dataset, which consists of face images and their corresponding ground-truth face shapes and camera parameters. Details on the construction of such training data are provided in Sec.~\ref{sec:data}. The DFF is trained using a convolutional neural network (CNN), which will be discussed in Sec.~\ref{sec:DFF}. Finally, we propose a new face alignment algorithm based on the DFF descriptor in Sec.~\ref{sec:Large_Method}. In the following, we first introduce the 3D face shape representation and the 3D Morphable Model, based on which our algorithm is developed.

\paraheading{3D Face Shape}
We represent a 3D face shape using a triangle mesh with $n$ vertices and fixed connectivity. Therefore, each face shape is determined by its vertex positions, represented as a $3 \times n$ matrix
\begin{equation}
\mathbf{S}= \left( \begin{matrix}
x_1 & x_2 & \cdots\ & x_n \\
y_1 & y_2 & \cdots\ & y_n \\
z_1 & z_2 & \cdots\ & z_n
\end{matrix} \right),
\label{eq:shaperepresentation}
\end{equation}
where each column vector corresponds to the 3D coordinates of a vertex. We denote a face image as $\mathbf{I}$, and assume the mapping from $\mathbf{S}$ to $\mathbf{I}$ to be a weak perspective projection with camera parameters $\mathbf{w} = (s, \alpha, \beta, \gamma, t_x, t_y)$, where $(t_x, t_y)$ represent the translation on the image plane, $s$ the scaling factor, $\alpha$ the pitch angle, $\beta$ the yaw angle, and $\gamma$ the roll angle. The three angles ($\alpha, \beta, \gamma$) determine a rotation matrix $\mathbf{R} \in \mathbb{R}^{3 \times 3}$. Then the 2D projection of a vertex $\mathbf{q} \in \mathbb{R}^3$ from the 3D face model can be written as:
\begin{equation}
P(\mathbf{q})
=
s
\left(
\begin{array}{ccc}
1 & 0 & 0\\
0 & 1 & 0\\
\end{array}
\right)
\mathbf{R}\mathbf{q}
+
\mathbf{t},
\label{eq:proj}
\end{equation}
where $\mathbf{t} = \left(\begin{matrix}t_x~~ t_y \end{matrix} \right)^T$.

\paraheading{Parametric Face Model}
The excessive degrees of freedom from vertex positions in Equation~\eqref{eq:shaperepresentation} can result in unnatural face shapes. From a perceptual point of view, realistic face shapes lie in a space of lower dimension~\cite{Meytlis2007}. Therefore, we follow the approach of 3D Morphable Model (3DMM)~\cite{blanz1999morphable} and represent a face shape and its albedo using a lower-dimensional linear model. We also introduce delta blendshapes to represent the deformation from a neutral face to its different expressions~\cite{Lewis2014}. The resulting parametric face model is
\begin{equation}
\begin{aligned}
\widehat{\mathbf{S}} = &~\overline{\mathbf{S}} + \mathbf{A}_{\textrm{id}} \mathbf{p}_{\textrm{id}} + \mathbf{A}_{\textrm{exp}} \mathbf{p}_{\textrm{exp}},\\
\widehat{\mathbf{T}} =&~\overline{\mathbf{T}} + \mathbf{B}_{\textrm{alb}} \mathbf{p}_{\textrm{alb}},
\end{aligned}
\label{eq:ParamFace}
\end{equation}
where $\widehat{\mathbf{S}} \in \mathbb{R}^{3n}$ stacks the vertex coordinates of $\mathbf{S}$, and $\widehat{\mathbf{T}} \in \mathbb{R}^{3n}$ represent the albedo values for the vertices. 
Here $\overline{\mathbf{S}} = \overline{\mathbf{S}}_{\textrm{id}} + \overline{\mathbf{S}}_{\textrm{exp}} \in \mathbf{R}^{3n}$ is the mean face shape, where $\overline{\mathbf{S}}_{\textrm{id}}$ and $\overline{\mathbf{S}}_{\textrm{exp}}$ are the mean identity and mean expression; 
$\overline{\mathbf{T}} \in \mathbf{R}^{3n}$ denotes the mean albedo values; 
$\mathbf{A}_{\textrm{id}}, \mathbf{B}_{\textrm{alb}} \in \mathbf{R}^{3n \times m_1}$ and $\mathbf{A}_{\textrm{exp}} \in \mathbb{R}^{3n \times m_2}$ are the bases for identity, albedo and expression, respectively, computed using PCA; $\mathbf{p}_{\textrm{id}}$, $\mathbf{p}_{\textrm{exp}}$, $\mathbf{p}_{\textrm{al}}$ are their linear combination coefficients. We denote the collection of shape parameters by $\mathbf{p}=(\mathbf{p}_{\textrm{id}}^T, \mathbf{p}_{\textrm{exp}}^T)^T$.
We choose $m_1 = 80$ and $m_2 = 79$ for our implementation.
The mean and basis identities are constructed from the Basel Face Model~\cite{paysan20093d}, while the mean and basis expressions are constructed using the FaceWarehouse dataset~\cite{cao2014facewarehouse}.

\begin{figure}[t]
	\centering
	%\fbox{\rule{0pt}{2in} \rule{.9\linewidth}{0pt}}
	\includegraphics[height=0.8in]{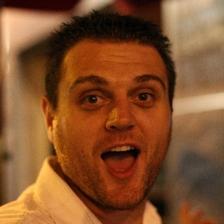}
	\includegraphics[height=0.8in]{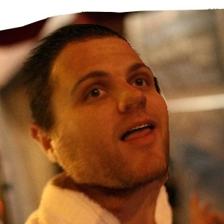}
	\includegraphics[height=0.8in]{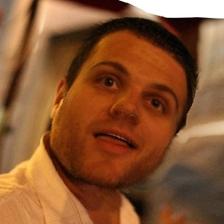}
	\includegraphics[height=0.8in]{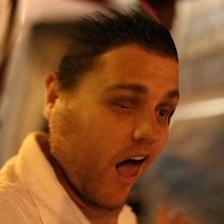}
	\caption{Examples of augmented training images. From the original face image (left), we construct a set of new images with different view directions and expressions.}
	\label{fig:FaceData}
\end{figure}

\subsection{Training Data Construction}
\label{sec:data}
To train the feature extraction model DFF, we need a large set of training data $\{(\mathbf{S}_i,\mathbf{I}_i,\mathbf{w}_i) \mid i=1,\cdots, N \}$, where $\mathbf{I}_i$ is the $i$-th face image, and $\mathbf{S}_i, \mathbf{w}_i$ are the corresponding ground-truth face shape and camera parameters. From $\mathbf{S}_i$ and $\mathbf{w}_i$, we can easily compute the location $\mathbf{U}_i$ and visibility $\mathbf{V}_i$ of ground-truth landmarks, by projecting the pre-selected landmark vertices from the face shape $\mathbf{S}_i$ onto the image plane and checking their visibility in 3D.

We first select $4308$ face images from the 300W dataset~\cite{sagonas2013300} and the Multi-PIE dataset~\cite{gross2010multi}, and then follow the approach of~\cite{guo20173d} to fit the parametric face model to each image and produce the shape, albedo, and camera parameters. Using these data, new face images are then generated by rendering the face shape with various expressions and camera parameters, resulting in a dataset that includes $80000$ face images with synthesized ground-truth face shapes and camera parameters. Each face image has a resolution of $224 \times 224$. Fig.~\ref{fig:FaceData} shows an example of the generated images.

%We keep the same train and test sets of divisions with 300W, and divide the whole Multi-PEI augmented images to train set.

\subsection{Deep Face Feature Training}
\label{sec:DFF}

\begin{figure}[t]
  \centering
\includegraphics[width=0.8\columnwidth]{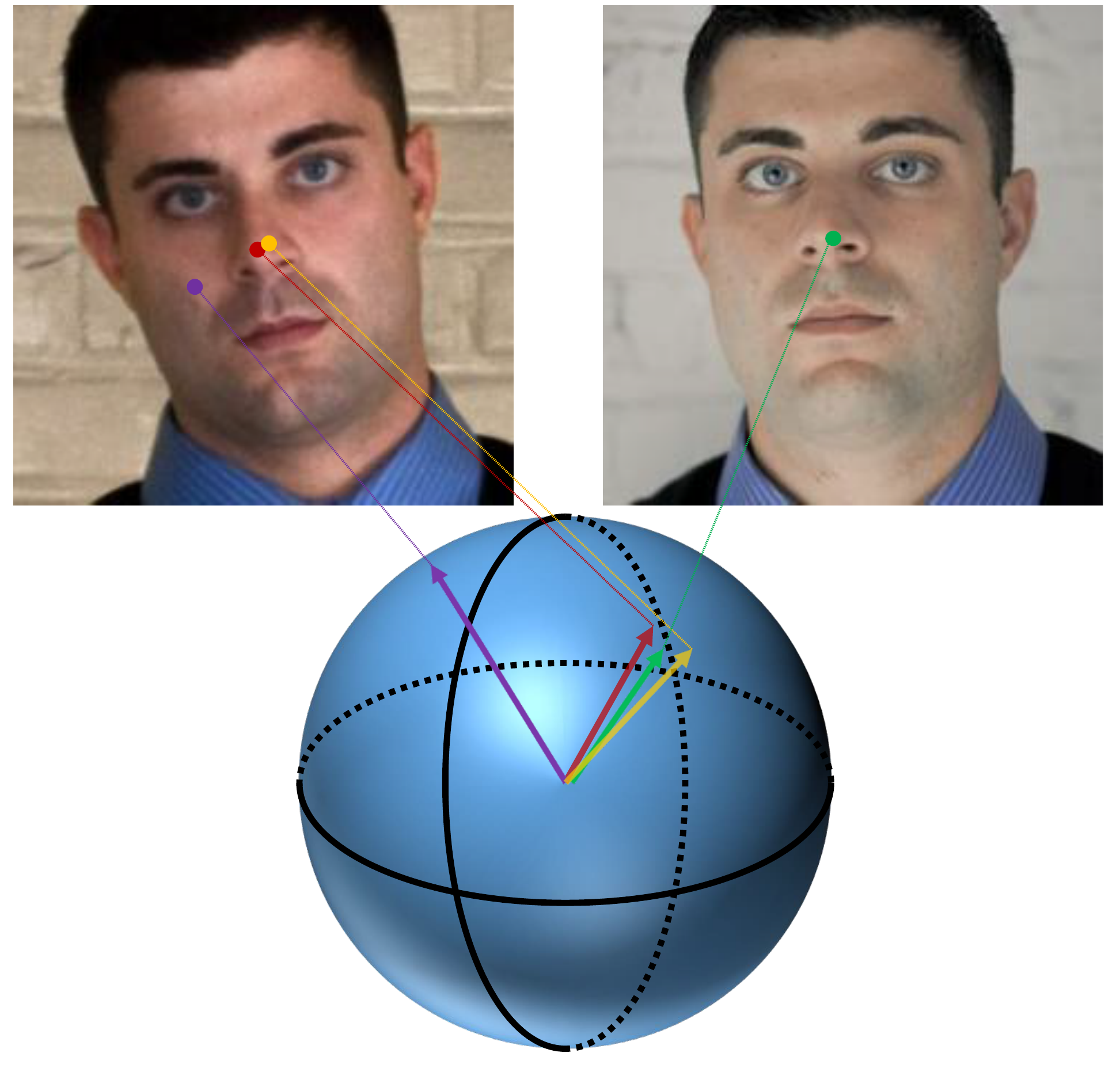}
  \caption{An illustration of the requirements for the DFF. If two pixels correspond to nearby points on the 3D face surface, their normalized DFF's should be close on the hypersphere; otherwise, they should be sufficiently far away from each other.}
  \label{fig:illustration}
\end{figure}

Many existing face alignment methods~\cite{xiong2013supervised,RenCWS16} rely on local features such as local binary feature (LBF) and SIFT. Although these methods work well in many scenarios, they might fail for input face images with large poses. One potential cause of the problem is that local features cannot utilize global information from the whole image and fail to recognize global structures such as specific face regions and self-occlusion. In this work, we propose a deep learning based end-to-end method, to extract for each face image pixel a feature vector that takes global information into account. Our method uses a neural network to map each pixel to a high-dimensional point, which is then normalized to have unit length. To effectively indicate and distinguish facial features, the normalized DFF descriptor should preserve the metric structure on the 3D face surface. In particular, for two pixels corresponding to the same anatomical region, their normalized DFF descriptors should be close to each other even if they are from different images with different poses, scales, and lighting conditions. On the other hand, for pixels corresponding to different facial parts, their normalized DFF descriptors should be sufficiently far away from each other even if their surrounding image regions have similar appearances. One such example is shown in Fig.~\ref{fig:illustration}.

To satisfy these criteria, we follow the approach of~\cite{wei2016dense} and train the DFF extractor to solve a series of classification problems. The approach is based on the following observation: if we randomly segment the face surface into uniform patches, then nearby points on the face surface are likely to lie on the same patch. Accordingly, if we perform classification of face image pixels into the segmentation patches according to their DFF descriptors, then pixels corresponding to nearby 3D face points should be classified into the same patch with high probability. In other words, the DFF extractor should lead to a small value of the classification loss function corresponding to the segmentation. To avoid bias towards a specific segmentation, we generate a large set of random segmentations for each face, and use the sum of their classification loss functions as the overall loss function for training the DFF extractor.

\begin{figure}[t]
	\centering
	%\fbox{\rule{0pt}{2in} \rule{.9\linewidth}{0pt}}
%	\includegraphics[width=0.32\columnwidth]{segs/seg_0.jpg}
%	\includegraphics[width=0.32\columnwidth]{segs/seg_1.jpg}
%	\includegraphics[width=0.32\columnwidth]{segs/seg_2.jpg}
	\includegraphics[width=0.9\columnwidth]{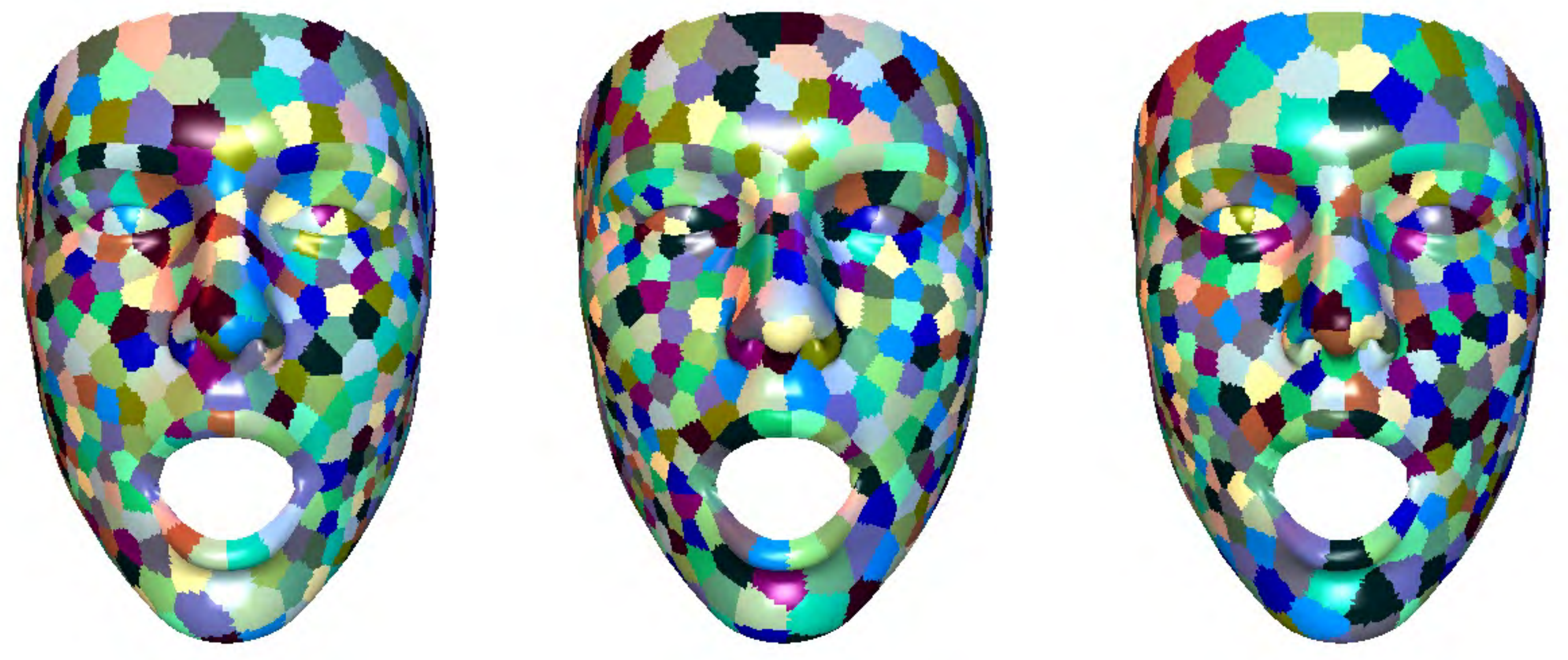}
	\caption{Examples of random segmentations of the 3D face surface into 500 patches.}
	\label{fig:segmentation}
\end{figure}

\begin{figure}[t]
	\centering
	\includegraphics[width=0.32\columnwidth]{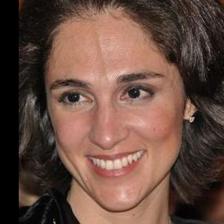}
	\includegraphics[width=0.32\columnwidth]{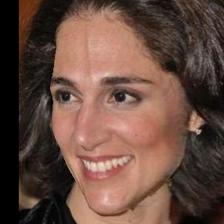}
	\includegraphics[width=0.32\columnwidth]{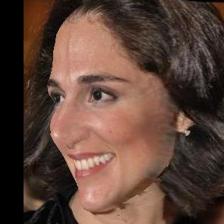}
	\includegraphics[width=0.32\columnwidth]{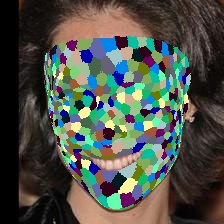}
	\includegraphics[width=0.32\columnwidth]{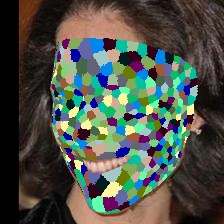}
	\includegraphics[width=0.32\columnwidth]{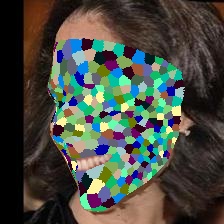}
	\caption{For each image corresponding to a segmented 3D face model, the visible patches are projected back onto the image plane to label the pixels they cover.}
	\label{fig:patch}
\end{figure}

In detail, we randomly generate 100 uniform segmentations of the mean face model, where each segmentation consists of 500 patches, and each set is a union of mesh faces (see Fig.~\ref{fig:segmentation} for examples). The segmentation is performed by computing centroidal Voronoi tessellation~\cite{Du1999} on the surface, using random sample points as initial generators.
Then for each ground-truth 3D face $\mathbf{S}_i$ in the training data set, we derive 100 segmentations by using the same set of faces for each patch as the mean face segmentation. For each segmentation, we project the visible patches to each training face image corresponding to $\mathbf{S}_i$, with the patch visibility determined from the camera parameters (see Fig.~\ref{fig:patch} for an example). The image pixels are then labeled according to the projected visible patches. We then use all the images with labels to train a CNN per-pixel classifier. The CNN consists of two parts. The first part is a DFF extractor that takes a $224 \times 224$ color image as input, and produces a 32-dimensional feature vector for each pixel. The architecture of our DFF extractor is similar to the u-net in~\cite{ronneberger2015u}, applying convolution and de-convolution layer symmetrically and concatenating with shallow feature map (see Fig.~\ref{fig:DFF_extractor}). The second part consists of 100 independent classification loss layers, one for each segmentation. Each layer takes the generated DFF's as input, and evaluates a classification loss function according to the segmentation. The sum of all loss functions is the final loss function for training the network. For each segmentation, we utilize the angular softmax~\cite{liu2017sphereface} with $m=1$ as the classification loss, which assumes a feature vector for each class such that input vectors are classified based on their angle to the class feature vectors. For each segmentation, we denote by $\mathbf{h}_j \in \mathbb{R}^{32}$ the unit feature vector for patch $j$ ($j=1,\ldots,500$), which is part of the parameters for the classification loss layer. Then its classification loss for an image is written as
\[
	l_k = \frac{1}{|\mathcal{P}|}\sum_{p \in \mathcal{P}} - \log\left(\frac{\exp(\mathbf{h}_{\tau(p)} \cdot \mathbf{f}_p)}{\sum_{j=1}^{500} \exp(\mathbf{h}_{j} \cdot \mathbf{f}_p )}\right),
\]
where $\mathcal{P}$ is the set of pixels corresponding to the visible patches, $\tau(p)$ is the index of the patch corresponding to pixel $p$, and $\mathbf{f}_p$ is the normalized DFF descriptor for $p$. The training optimizes the parameters of the DFF extractor as well as the classification loss layers, to reduce the sum of all classification loss.

\begin{figure*}
	\centering
		%\fbox{\rule{0pt}{2in} \rule{.9\linewidth}{0pt}}
		\includegraphics[width=\textwidth]{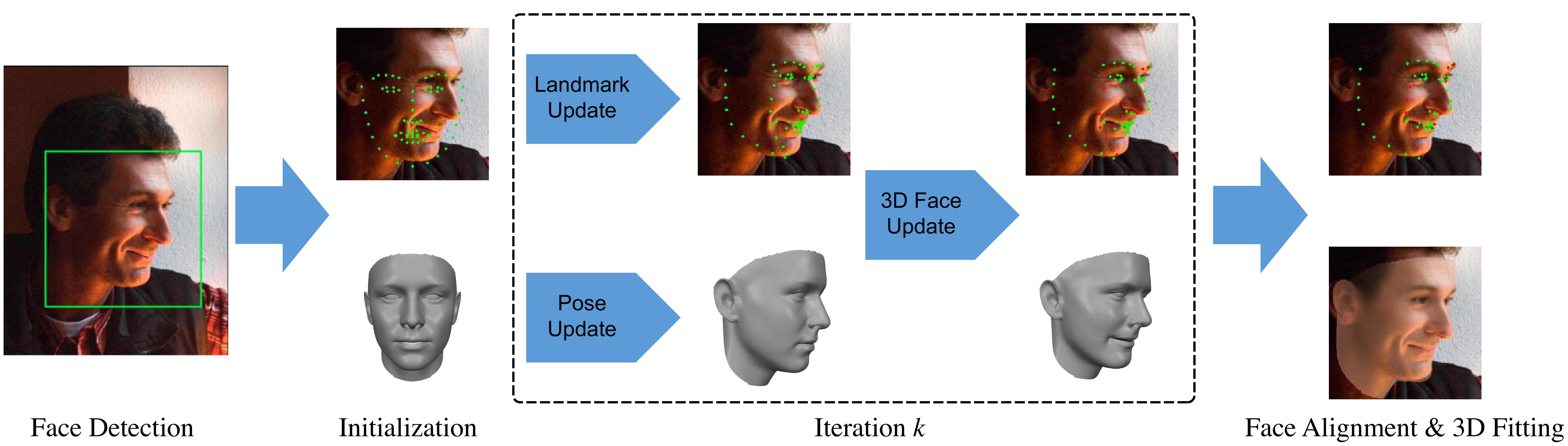}
	\caption{An overview of our face alignment algorithm pipeline. The input is a face image with a detected face region. Starting from the initial 3D face model and its projected 2D landmark locations (shown in green), we evaluate the DFF descriptors at the projected landmarks, and use them to compute the target landmark locations and update the camera parameters according to a generic descent direction learned from the training data. Afterwards, we update the 3D face model to align with the target landmark locations, and recompute its projected 2D landmarks and their visibility (with invisible landmarks shown in red). This process is iterated until convergence.}
	\label{fig:overview}
\end{figure*}

\subsection{Large Pose Face Alignment}
\label{sec:Large_Method}
Classical cascaded regression algorithms such as the Supervised Descent Method (SDM)~\cite{xiong2013supervised} use SIFT descriptors around the landmarks to regress their locations. These methods achieve state-of-the-art performance for small- and medium-pose face alignment. However, their results are not satisfactory for large poses or more complicated scene conditions. This is because for general descriptors such as SIFT, their capability of identifying key points can decrease drastically under large pose variation. Moreover, without visibility information, SIFT descriptors for landmarks that are occluded in some images may interfere with the regression.

To address these issues, we propose an iterative approach that utilizes the DFF descriptor instead of SIFT, and uses the parametric face model to estimate face pose and landmark visibility. Given an input face image, we find a 3D parametric face model as well as the camera parameters, such that the projected 3D landmarks according to the camera parameters align with the face image.
Starting from the initial parametric face model and the initial camera parameters, we first project the visible 3D landmark vertices onto the image plane to obtain 2D landmark locations, and evaluate the DFF descriptors of the face image at these landmark locations. From the DFF descriptors, we determine the target new locations of 2D landmarks and update the camera parameters, using a learned mapping from DFF descriptors to landmark and camera parameter updates that improve their accuracy. Afterwards, the parametric 3D face model is updated accordingly, such that its projected landmark locations using the new camera parameters are as close as possible to the target locations. This process is iterated until convergence. Figure~\ref{fig:overview} illustrates the pipeline of our algorithm. In the following, we explain each step in detail.

%In each iteration, we compute the DFF descriptors of the  at the projected 3D 
%DFF descriptors are used to regress landmark locations and camera parameters, followed by a refinement step that updates the landmarks visibility and location. With more DFF descriptors as input in each regression iteration, we can achieve high accuracy and more robust results. In our implementation, the number of face feature landmarks set and landmarks set used for DFF extraction are denoted as $J$ and $U$, and are respectively setted with $68$ and $160$. The first set including $68$ landmarks chosen according to popular Multi-PIE~\cite{gross2010multi} $68$ points make-up. The second set is a uniform dense sampling on 3DMM model based on the first set. Fig.~\ref{fig:Landmarks} shows the two landmark sets. The projected 2D coordinates of these landmarks are respectively represented as $\mathbf{X}\in\mathbb{R}^{2\times J}$ and $\mathbf{U} \in \mathbb{R}^{2 \times L}$.

\begin{figure}[b!]
\centering
%\fbox{\rule{0pt}{2in} \rule{.9\linewidth}{0pt}}
\includegraphics[totalheight=1.5in]{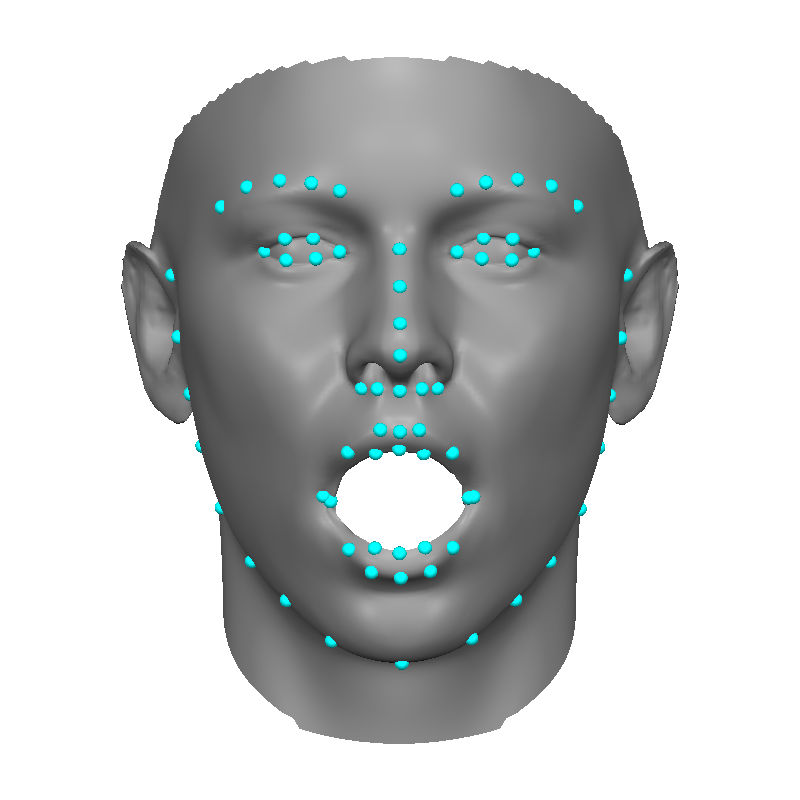}
\includegraphics[totalheight=1.5in]{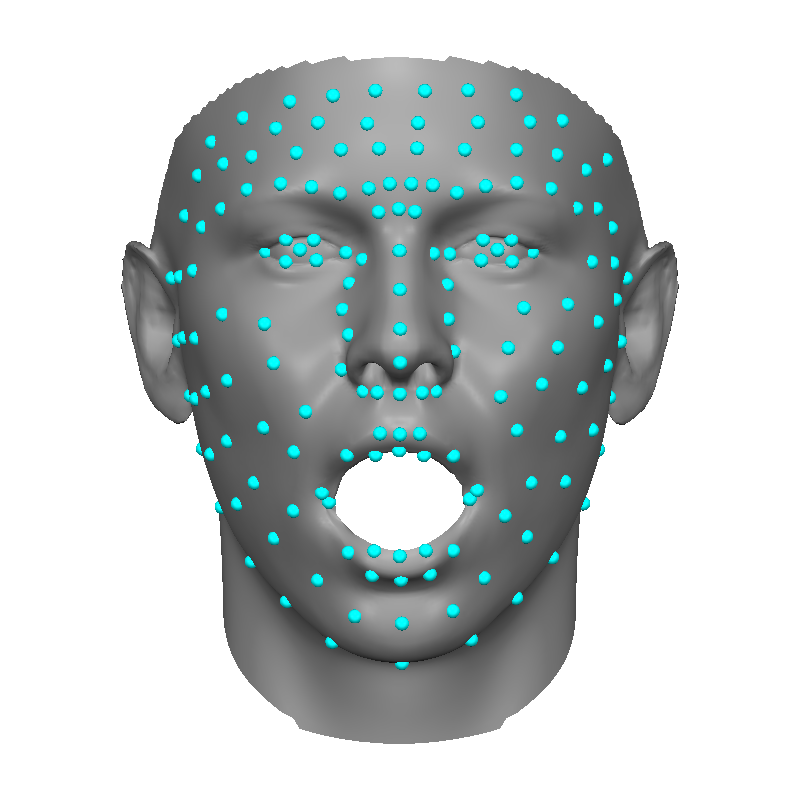}
\caption{We perform face alignment using 68 landmarks (shown on the left), which are chosen on the 3D face mesh according to the 68 feature points of the Multi-PIE dataset~\cite{gross2010multi}. In our face alignment algorithm, the update of landmarks and camera parameters is driven by DFF descriptors corresponding to $160$ landmarks on the 3D face model (shown on the right), which is a superset of the 68 landmarks.}
\label{fig:Landmarks}
\end{figure}

\subsubsection{Initialization}
\label{sec:Initialization}
Given an input face image, we first run a face detector to locate the face region. We initialize the parametric face model $\mathbf{S}^{(0)}$ and the camera parameters $\mathbf{w}^{(0)}$ to the mean face shape and mean camera parameters, respectively. The landmark vertices on $\mathbf{S}^{(0)}$ are projected onto the image plane according to Equation~\eqref{eq:proj} using camera parameters $\mathbf{w}^{(0)}$, to determine the initial 2D landmark locations as well as their visibility. In this work, the alignment between the 3D face model and the 2D face image is done via 68 landmark vertices, chosen on the 3D face mesh to match the 68 landmark points used in the Multi-PIE data set~\cite{gross2010multi} (see Fig.~\ref{fig:Landmarks} left). The initial 2D locations of these landmarks are denoted by $\mathbf{X}^{(0)} \in \mathbb{R}^{136}$. In addition, we use a denser set of landmark vertices to evaluate DFF descriptors, in order to capture more global information of the face shape and produce more robust results. Specifically, we densely sample the face mesh to obtain 160 landmark vertices, which includes the previous 68 landmarks (see Fig.~\ref{fig:Landmarks} right). The initial 2D locations of these landmarks are denoted by $\mathbf{U}^{(0)} \in \mathbb{R}^{320}$, and their visibility is indicated using a binary vector $\mathbf{V}^{(0)} \in \mathbb{R}^{160}$.

%and compute its The face shape $\mathbf{S}_i$ is initialized by the mean shape $\bar{\mathbf{S}}$, and the landmarks $\mathbf{U}_i$, $\mathbf{X}_i$ and visibility $\mathbf{V}_i$ are initialized by projecting $\bar{\mathbf{S}}$ to corresponding image with mean camera parameters $\mathbf{w}_0$. In the $k$-th iteration, DFF $f(\mathbf{I}_i,\mathbf{U}_i^k,\mathbf{V}_i^k)$ is extracted from image $\mathbf{I}_i$ based on current landmark locations $\mathbf{U}_i^k$ and visibility $\mathbf{V}_i^k$. The feature vector $f$ is a concatenation of the DFF descriptors around $\mathbf{U}_i$ on $\mathbf{I}_i$, which means that $f$ is a $32L$-dimensional vector. If a landmark is invisible, its corresponding entries in $f$ will be set to be zero. Then, $f$ is used to regress $\mathbf{X}_i$ and $\mathbf{w}_i$ respectively. Based on the updated $(\mathbf{X}_i, \mathbf{w}_i)$, we iteratively optimize face shape $\mathbf{S}_i$, and then refine $(\mathbf{U}_i,\mathbf{X}_i,\mathbf{V}_i)$ until convergence. The algorithm details of each componenet are given in the following.

\subsubsection{Updating Landmarks and Camera Parameters} 
\label{sec:LandmarkCameraParamUpdate}
In the $k$-th iteration, we first evaluate the DFF descriptors of the dense landmark set using their current locations $\mathbf{U}^{(k)}$ and visibility $\mathbf{V}^{(k)}$. The DFF descriptors are concatenated into a vector $\mathbf{F}^{(k)} \in \mathbb{R}^{5120}$. For invisible landmarks, their corresponding components $\mathbf{F}^{(k)}$ are set to zero. Using $\mathbf{F}^{(k)}$, we compute new camera parameters $\mathbf{w}^{(k+1)}$ as well as target locations for new 2D landmarks $\widehat{\mathbf{X}}^{(k+1)}$, both of which should improve the accuracy of alignment. We adapt the approach of~\cite{xiong2013supervised} to determine $\mathbf{w}^{(k+1)}$ and $\widehat{\mathbf{X}}^{(k+1)}$, by computing the displacements $\mathbf{w}^{(k+1)} - \mathbf{w}^{(k)}$ and $\widehat{\mathbf{X}}^{(k+1)} - \mathbf{X}^{(k)}$ as a linear function of the DFF descriptors $\mathbf{F}^{(k)}$:
\begin{align}
	\widehat{\mathbf{X}}^{(k+1)} &=  \mathbf{X}^{(k)} + \mathbf{R}_{\mathbf{X}}^{(k)} \mathbf{F}^{(k)} + \mathbf{b}_{\mathbf{X}}^{(k)},\label{eq:LandmarkUpdate}\\
	{\mathbf{w}}^{(k+1)} &=  \mathbf{w}^{(k)} + \mathbf{R}_{\mathbf{w}}^{(k)} \mathbf{F}^{(k)} + \mathbf{b}_{\mathbf{w}}^{(k)}.\label{eq:CamParmUpdate}
\end{align}
Here matrix $\mathbf{R}_{\mathbf{X}}^{(k)}$ and vector $\mathbf{b}_{\mathbf{X}}^{(k)}$ represent a \emph{generic descent direction} that improves the accuracy of landmark locations according to the current DFF descriptors~\cite{xiong2013supervised}. They are learned from the training data set constructed in Sec.~\ref{sec:data}, by solving a sequence of regression problems. Similarly, matrix $\mathbf{R}_{\mathbf{w}}^{(k)}$ and vector $\mathbf{b}_{\mathbf{w}}^{(k)}$ are learned from the training data set. The details of learning the generic descent directions are explained later in Sec.~\ref{eq:GenericDescent}.

\subsubsection{Updating 3D Face Model} 
\label{sec:3DFaceUpdate}
Since the target 2D landmark positions $\widehat{\mathbf{X}}^{(k+1)}$ are determined by simply applying a generic descent step, they do not necessarily correspond to a specific parametric 3D face model. Therefore, to ensure the validity of landmark locations, we compute an updated parametric face model $\mathbf{S}^{(k+1)}$ according to $\widehat{\mathbf{X}}^{(k+1)}$, and determine new 2D landmark locations by projecting the updated face model according to the camera parameters ${\mathbf{w}}^{(k+1)}$. This results in an optimization problem for the shape parameters $\mathbf{p} = (\mathbf{p}_{\textrm{id}}^T, \mathbf{p}_{\textrm{exp}}^T)^T$ that determine the face model $\mathbf{S}$ according to Equation~\eqref{eq:ParamFace}: 
\begin{equation}
	\mathbf{p}^{(k+1)} = ~\argmin_{\mathbf{p}}~\omega_{\textrm{lan}} \emph{E}_{\textrm{lan}}^{(k)}(\mathbf{p}) + \omega_{\textrm{reg}} \emph{E}_{\textrm{reg}}^{(k)}(\mathbf{p}).
\label{eq:shape}
\end{equation}
Here $\emph{E}_{\textrm{lan}}^{(k)}$ is a landmark fitting term, $\emph{E}_{\textrm{reg}}^{(k)}$ is a regularization term, and $\omega_{\textrm{lan}}, \omega_{\textrm{reg}}$ are their weights. The terms are defined as:
\begin{align}
\emph{E}_{\textrm{lan}}^{(k)}(\mathbf{p}) &= \|\widehat{\mathbf{X}}^{(k+1)} - \mathbf{Y}(\mathbf{w}^{(k+1)}, \mathbf{p})\|^2,\\
\emph{E}_{\textrm{reg}}^{(k)}(\mathbf{p}) &= \mathbf{p}_{\textrm{id}}^T~\mathbf{D}_{\textrm{id}}^{-1}~\mathbf{p}_{\textrm{id}}
	+ \mathbf{p}_{\textrm{exp}}^T~\mathbf{D}_{\textrm{exp}}^{-1}~\mathbf{p}_{\textrm{exp}}.
\end{align}
Here $\mathbf{Y}(\mathbf{w}^{(k+1)}, \mathbf{p})$ is a vector storing the projected 2D coordinates of the 68 landmarks from the face model with shape parameters $\mathbf{p}$, according to camera parameters $\mathbf{w}^{(k+1)}$. $\mathbf{D}_{\textrm{id}}$ and $\mathbf{D}_{\textrm{exp}}$ are diagonal matrices that store the eigenvalues of the identity and expression covariance matrices corresponding to bases $\mathbf{A}_{\textrm{id}}$ and $\mathbf{A}_{\textrm{exp}}$ in Equation~\eqref{eq:ParamFace}. This is a linear least-squares problem and can be solved efficiently.

After solving for the shape parameters $\mathbf{p}^{(k+1)}$, we determine the new 3D face shape $\mathbf{S}^{(k+1)}$, and project its landmark vertices to obtain the new 2D landmark coordinates $\mathbf{U}^{(k+1)}, \mathbf{X}^{(k+1)}$, as well as the visibility $\mathbf{V}^{(k+1)}$ for $\mathbf{U}^{(k+1)}$. $\mathbf{V}^{(k+1)}$ can be easily determined on the GPU by rendering the triangles of the 3D face mesh and the landmark vertices to the OpenGL depth buffer, with the OpenGL projection matrix determined from the corresponding camera parameters.

We iteratively apply landmark update, camera parameters update and 3D face update until convergence. In our experiments, three iterations are sufficient for good results.

\subsubsection{Learning the Generic Descent Directions}
\label{eq:GenericDescent}
In Sec.~\ref{sec:LandmarkCameraParamUpdate}, the computation of target landmark positions $\widehat{\mathbf{X}}^{(k+1)}$ and new camera parameters ${\mathbf{w}}^{(k+1)}$ require the generic descent directions --- represented using matrices $\mathbf{R}_{\mathbf{X}}^{(k)}, \mathbf{R}_{\mathbf{w}}^{(k)}$ and vectors $\mathbf{b}_{\mathbf{X}}^{(k)}, \mathbf{b}_{\mathbf{w}}^{(k)}$ --- that map the current DFF descriptors to the displacements from landmark positions $\mathbf{X}^{(k)}$ and camera parameters $\mathbf{w}^{(k)}$, respectively. 
Following~\cite{xiong2013supervised}, we learn the generic descent directions in each iteration using the training images $\mathbf{I}_i$ ($i=1, \ldots, N$) together with their ground-truth landmark locations $\mathbf{X}_i^\ast$ and camera parameters $\mathbf{w}_i^\ast$. Using a generic face model $\overline{\mathbf{S}}$ for all the training images, we iteratively perform landmark update, camera parameters update and 3D face update similar to Secs.~\ref{sec:Initialization} to \ref{sec:3DFaceUpdate}, and determine the generic descent directions via linear regression.

In detail, we initialize the $\overline{\mathbf{S}}^{(0)}$ to the mean face shape, and initialize the camera parameters $\mathbf{w}_i^{(0)}$, landmark locations $\mathbf{X}_i^{(0)}, \mathbf{U}_i^{(0)}$ and visibility $\mathbf{V}_i^{(0)}$ for image $\mathbf{I}_i$ as described in Sec.~\ref{sec:Initialization}. In the $k$-th iteration, we evaluate the DFF descriptors $\mathbf{F}_i^{(k)}$ of image $\mathbf{I}_i$ according to its landmark locations $\mathbf{U}_i^{(k)}$ and visibility $\mathbf{V}_i^{(k)}$, as described in Sec.~\eqref{sec:LandmarkCameraParamUpdate}. The generic descent direction $(\mathbf{R}_{\mathbf{X}}^{(k)}, \mathbf{b}_{\mathbf{X}}^{(k)})$ for landmark locations in iteration $k$ is then determined such that the resulting updated landmarks are as close as possible to the ground-truth landmarks $\mathbf{X}_i^{\ast}$ for all the training images. We thus compute $(\mathbf{R}_{\mathbf{X}}^{(k)}, \mathbf{b}_{\mathbf{X}}^{(k)})$ by solving a regression problem
\begin{align}
	(\mathbf{R}_{\mathbf{X}}^{(k)}, \mathbf{b}_{\mathbf{X}}^{(k)}) =  \min_{\mathbf{R}, \mathbf{b}} & \sum_{i=1}^N \| \mathbf{X}_i^\ast - \mathbf{X}_i^{(k)} - \mathbf{R} \mathbf{F}_i^{(k)} - \mathbf{b} \|_2^2 \nonumber\\
	& + \lambda_1 \left(\|\mathbf{R}\|_F^2 + \|\mathbf{b}\|_2^2\right),
\end{align}
where the second term is a regularization, and $\lambda_1$ is a positive weight. Similarly, the generic descent direction $(\mathbf{R}_{\mathbf{w}}^{(k)}, \mathbf{b}_{\mathbf{w}}^{(k)})$ is determined via a regression problem
\begin{align}
(\mathbf{R}_{\mathbf{w}}^{(k)}, \mathbf{b}_{\mathbf{w}}^{(k)}) =  \min_{\mathbf{R}, \mathbf{b}} & \sum_{i=1}^N \| \mathbf{w}_i^\ast - \mathbf{w}_i^{(k)} - \mathbf{R} \mathbf{F}_i^{(k)} - \mathbf{b} \|_2^2 \nonumber\\
& + \lambda_2 \left(\|\mathbf{R}\|_F^2 + \|\mathbf{b}\|_2^2\right).
\end{align}
Both regressions are linear least-squares problems and can be solved efficiently. Afterwards, we compute the target landmark locations $\widehat{\mathbf{X}}_i^{(k+1)}$ and updated camera parameters ${\mathbf{w}}_i^{(k+1)}$ for each training image according to Equations~\eqref{eq:LandmarkUpdate} and \eqref{eq:CamParmUpdate}. Then the generic face model $\overline{\mathbf{S}}$ is updated by optimizing its shape parameters $\overline{\mathbf{p}} = (\overline{\mathbf{p}}_{\textrm{id}}^T, \overline{\mathbf{p}}_{\textrm{exp}}^T)^T$ to align with the target landmark locations in all training images, similar to Equation~\eqref{eq:shape}. To be precise, we solve for
\begin{align}
	\overline{\mathbf{p}}^{(k+1)} = & \argmin_{\overline{\mathbf{p}}} \frac{\omega_{\textrm{lan}}}{N} \sum_{i=1}^N \|\widehat{\mathbf{X}}_i^{(k+1)} - \mathbf{Y}(\mathbf{w}_i^{(k+1)}, \overline{\mathbf{p}})\|^2 \nonumber\\
	& + \omega_{\textrm{reg}} \left(\overline{\mathbf{p}}_{\textrm{id}}^T~\mathbf{D}_{\textrm{id}}^{-1}~\overline{\mathbf{p}}_{\textrm{id}}
	+ \overline{\mathbf{p}}_{\textrm{exp}}^T~\mathbf{D}_{\textrm{exp}}^{-1}~\overline{\mathbf{p}}_{\textrm{exp}}\right).
\end{align}
Using $\overline{\mathbf{p}}^{(k+1)}$, we derive the new generic face model $\overline{\mathbf{S}}^{(k+1)}$, and compute the new landmark locations $\mathbf{X}_i^{(k+1)}, \mathbf{U}_i^{(k+1)}$ and visibility $\mathbf{V}_i^{(k+1)}$ for each training image. We repeat the above procedures until the generic descent directions are learned for all the required iterations.

\section{Experiments}
In this section, we evaluate the effectiveness of our approach by applying it to feature point matching and face alignment, and comparing the results with other existing approaches. 

%conduct extensive experiments to demonstrate the effectiveness of the proposed approach. To show the effectiveness of the trained DFF, feature points matching results on multi-view face images from the stereo face dataset~\cite{fransens2005parametric} will be given in Sec.~\ref{sec:performance}. Then, the performance between our method and other existing approaches for face alignment in the wild is compared, including large-pose face alignment and medium pose face alignment.  

%\comment{Because our train loss doesn't control DFF descriptor's property for background pixels, we need to exclude background interference when evaluating DFF performance for sparse or dense matching. We use our train data to train a simple binary classifier to extract face region of face image.}

\subsection{Feature Matching}
\label{sec:performance}
We first evaluate the effectiveness of DFF in capturing global structural information and identifying facial features.  
Since SIFT is widely used for feature matching and landmark regression, we compare the performance of DFF and SIFT in matching features of multi-view face images. For SIFT matching, we follow the approach of~\cite{lowe2004distinctive}. Specifically, given the source image $\overline{\mathbf{I}}_1$ and the target image $\overline{\mathbf{I}}_2$, we first identify for each image a set of feature points as describe in~\cite{lowe2004distinctive}. We denote the two feature point sets by $S_1$ and $S_2$, respectively. For each point $p_1$ in the set $S_1$, we find a point $p_2$ from the set $S_2$ whose SIFT descriptor is the closest to that of $p_1$. Following~\cite{lowe2004distinctive}, we consider the pair $(p_1, p_2)$ is as valid matching only if the ratio $d_{1,2}/d'_{1,2}$ is smaller than a certain threshold, where $d_{1,2}, d'_{1,2}$ are the SIFT descriptor angles between $p_1, p_2$ and between $p_1, p'_2$ respectively, with $p'_{2}$ a point from $S_2$ whose SIFT descriptor is the second closest to that of $p_1$. In our experiments, the ratio threshold is fine-tuned to $1/1.3$ to achieve the best results. 

For DFF matching, we first evaluate the DFF descriptor for each feature point from $S_1$ that lies on the face region, as well as DFF descriptors for all the face pixels in image $\overline{\mathbf{I}}_2$. Then for each face feature point $q_1$ from $S_1$, we find a face point $q_2$ from $\overline{\mathbf{I}}_2$ whose DFF descriptor vector has the smallest angle from the DFF descriptor of $q_1$. For accurate matching, the pair is considered valid only if the angle between their descriptor vectors is less than a certain threshold. We set the threshold to $30^{\circ}$ in our experiments, to trade off between accuracy and the number of valid pairs. 

\begin{figure}[!t]
	\centering
\includegraphics[width=0.49\columnwidth]{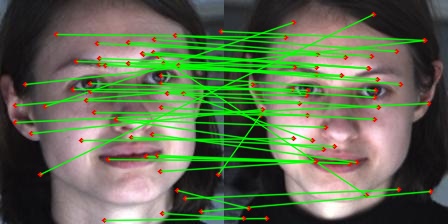}
\includegraphics[width=0.49\columnwidth]{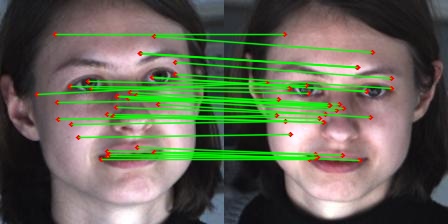}\\

\includegraphics[width=0.49\columnwidth]{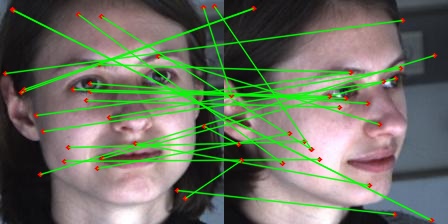}
\includegraphics[width=0.49\columnwidth]{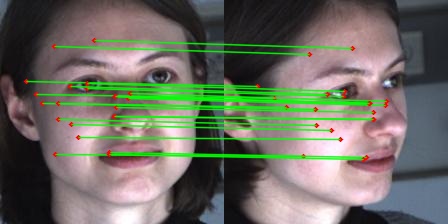}

\includegraphics[width=0.49\columnwidth]{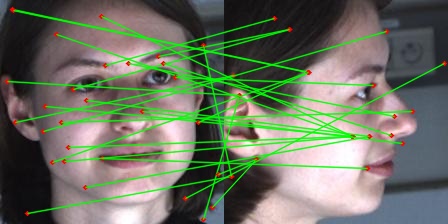}
\includegraphics[width=0.49\columnwidth]{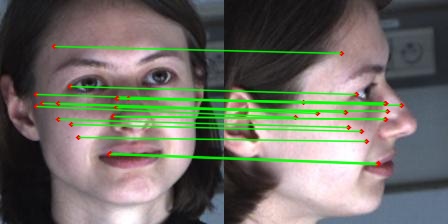}
	\caption{Comparison of feature matching results using SIFT (left) and our DFF descriptor (right), on three image pairs that share the same target image and use different target images with small, medium, and large view angle differences.}
	\label{fig:sparse_matching}
\end{figure}

Fig.~\ref{fig:sparse_matching} compares the results using the two approaches, on three image pairs with the same source image and different target images with small, medium, and large view angle difference, respectively. It can be seen that DFF leads to more stable and accurate than SIFT, especially for the image pair with large view angle difference. Moreover, since the DFF descriptors encode global structural information, the DFF matching results are consistent across different image pairs, with the same source feature point mapped to consistent points in different target images. This is not the case for SIFT, as it only considers local features. This is apparent on the image pair with large view angle difference, where DFF matching excludes source feature points that corresponding to invisible regions in the target image, while SIFT matches them to the other side of the face due to local structural similarity.

\begin{figure}[!t]
\centering
%\fbox{\rule{0pt}{2in} \rule{.9\linewidth}{0pt}}
\includegraphics[width=\columnwidth]{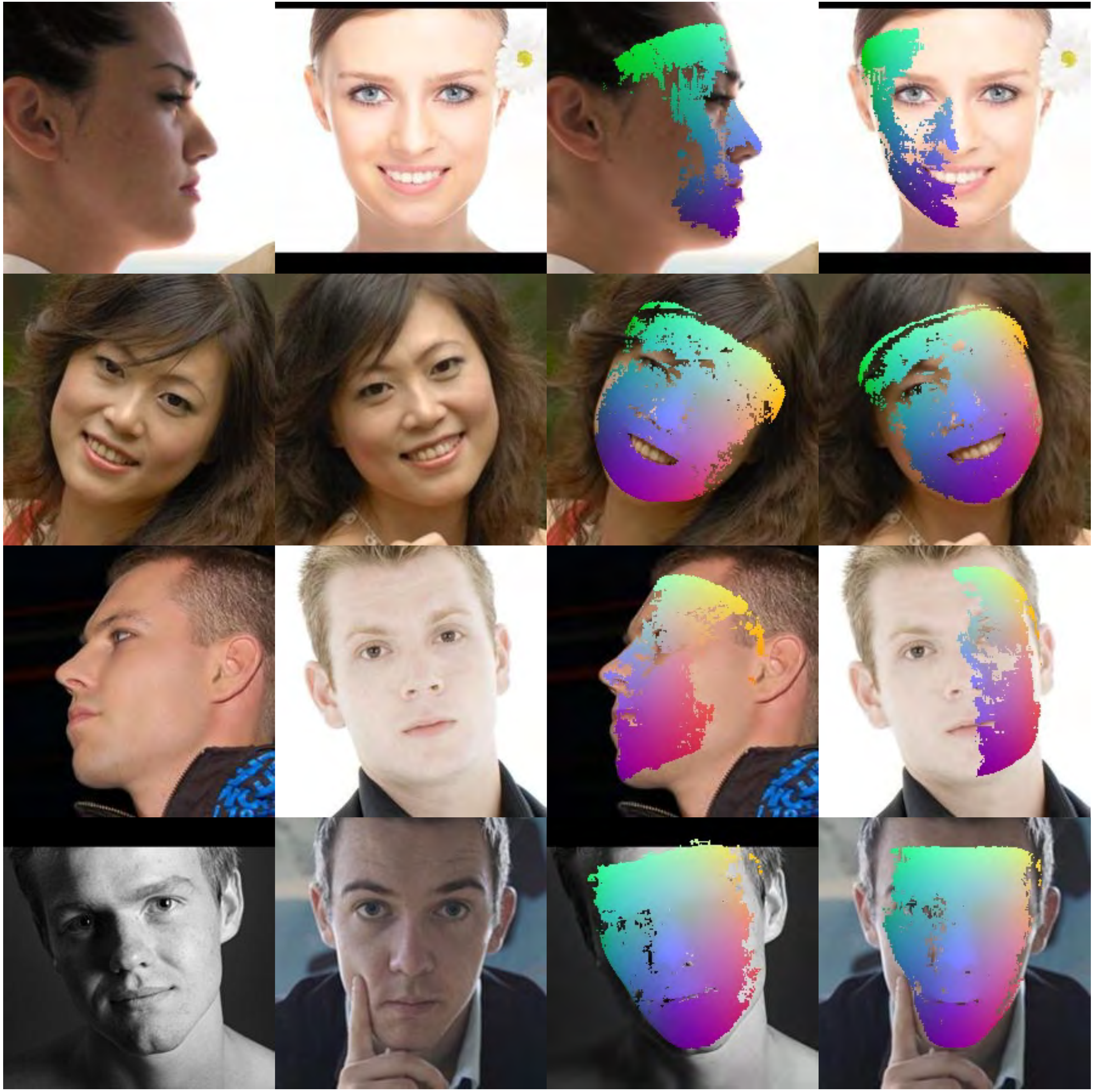}
\caption{Dense correspondence between multi-view face images based on the DFF descriptor. Each row shows the source image and the target image, and visualizes the correspondence between their face pixels via color coding.}
\label{fig:dense_match}
\end{figure}

We also apply DFF for dense matching between two face images with different poses, using images from the AFLW2000-3D dataset~\cite{zhu2016face}. We evaluate the DFF descriptor for each face pixel in the source image as well as the target image. Then each source face pixel is matched to a face pixel in the target image whose DFF descriptor has the smallest angle from the source DFF descriptor. The matching is considered as valid if the angle is smaller than $12^{\circ}$. Here we use a smaller threshold for valid matching, in order to reduce ambiguity due to the large number of source and target pixels. Fig.~\ref{fig:dense_match} shows some dense matching results, using color coding to indicate corresponding pixels.  It shows excellent performance of DFF for dense matching between face images with very different poses.

\subsection{Face Alignment}
\label{sec:large-pose comparison}

\begin{figure}[t]
	\setlength{\lineskip}{0pt}
	\centering
	%\fbox{\rule{0pt}{2in} \rule{.9\linewidth}{0pt}}
	\includegraphics[totalheight=1.0in]{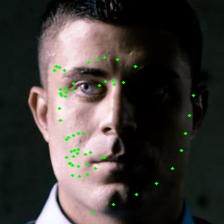}
	\includegraphics[totalheight=1.0in]{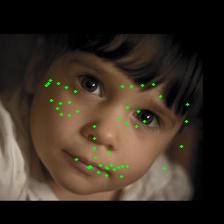}
	\includegraphics[totalheight=1.0in]{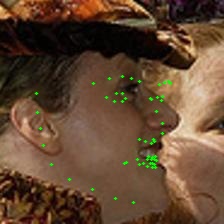}
	\includegraphics[totalheight=1.0in]{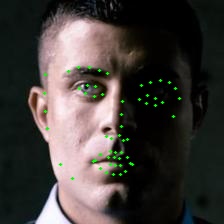}
	\includegraphics[totalheight=1.0in]{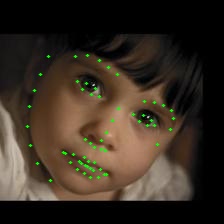}
	\includegraphics[totalheight=1.0in]{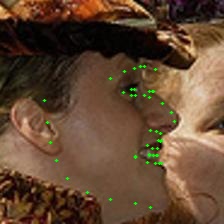}
	\caption{Face alignment results from the method described in Sec.~\ref{sec:Large_Method}, using SIFT (top row) and DFF (bottom row) as the feature descriptor, respectively. Due to the global information encoded in DFF, it leads to more robust and accurate results than SIFT.}
	\label{fig:dff_sift_compare}
\end{figure}

In Figure~\ref{fig:dff_sift_compare}, we evaluate the robustness of our face alignment algorithm, by applying it to the face images with large area of self shadow and with large face pose. For comparison, we also apply a method similar to the one in Sec.~\ref{sec:Large_Method}, using SIFT instead of DFF as the feature descriptors. Thanks to global information encoded in DFF, it produces more robust and accurate results than SIFT on these challenging examples.

To evaluate the performance of our approach for large pose face alignment, we test it using face images in the wild from the AFLW dataset\footnote{https://www.tugraz.at/institute/icg/research/team-bischof/lrs/downloads/aflw/} and the AFLW2000-3D dataset\footnote{http://www.cbsr.ia.ac.cn/users/xiangyuzhu/projects/3DDFA/main.htm}. The AFLW dataset contains face images with 21 visible ground truth landmarks, while the AFLW2000-3D dataset consists of fitted 3D faces for the first 2000 AFLW samples and can be used for 3D face alignment evaluation. In our experiments, the alignment accuracy is measured by the Normalized Mean Error (NME), which is the average of landmark error normalized by the bounding box size~\cite{jourabloo2015pose}. For AFLW, we compute NME using only the visible landmarks and the bounding boxes provided in the dataset. For AFLW2000-3D, the bounding box that encloses all the 68 ground truth landmarks are used to compute NME, similar to~\cite{zhu2016face}. Since the results reported in~\cite{zhu2016face} are obtained using a model trained with the 300W-LP dataset, we also learn the generic descent directions using the 300W-LP dataset for consistency. 

All $24384$ face images from the AFLW dataset, with yaw angles ranging from $-90^{\circ}$ to $90^{\circ}$, are used for testing. To show the robustness of our method for large yaw angles, we divide the images into 3 subsets according to their absolute yaw angles: $[0^{\circ},30^{\circ}]$,$[30^{\circ}, 60^{\circ}]$ and $[60^{\circ}, 90^{\circ}]$, with $14032$, $5949$ and $4403$ images respectively. For the AFLW2000-3D dataset, we follow the same experimental setting as~\cite{zhu2016face}. Tables~\ref{tab:AFLW} and \ref{tab:AFLW2000} compare the results using our approach and other existing alignment methods on the two datasets, with the best results highlighted in bold font.
The results of other methods are gathered from~\cite{zhu2016face} and ~\cite{yu2017learning}.
Our method significantly outperforms RCPR~\cite{burgos2013robust}, ESR~\cite{cao2014face}, SDM~\cite{xiong2013supervised} and Yu et al.~\cite{yu2017learning}, especially for samples within $[60^{\circ}, 90^{\circ}]$ yaw angles. Our method achieves similar performance with 3DDFA+SDM (which applies SDM to refine the results from 3DDFA~\cite{zhu2016face}) for $[0,60^{\circ}]$ yaw angles, and better results for $[60^{\circ},90^{\circ}]$ yaw angles.

Yu et al.~\cite{yu2017learning} observed that a decent-quality fitting can have a high NME due to the
subjective nature of the contour and invisible landmarks in some results of AFLW2000-3D. To better understand the performance of their method, they exclude the contour and invisible landmarks and evaluate the NME using only the inner and visible landmarks. Tab.~\ref{tab:AFLW2000_visi} shows the comparison result with the same setting, where our method achieves the best results.

\begin{table}[!b]
\caption{The NME ($\%$) on AFLW Dataset (21 Points)}
\label{tab:AFLW}
\begin{center}
\begin{tabular}{|c||c|c|c|c|c|}
\hline
Method &[0,30]&[30,60]&[60,90]&Mean&Std\\
\hline
RCPR~\cite{burgos2013robust} (300W-LP)  &$5.43$&$6.58$&$11.53$&$7.85$&$3.24$\\
\hline
ESR~\cite{cao2014face} (300W-LP)  &$5.66$&$7.12$&$11.94$&$8.24$&$3.29$\\
\hline
SDM~\cite{xiong2013supervised} (300W-LP)  &$4.75$&$5.55$&$9.34$&$6.55$&$2.45$\\
\hline
3DDFA~\cite{zhu2016face}  &$5.00$&$5.06$&$6.74$&$5.60$&$0.99$\\
\hline
3DDFA+SDM  &$4.75$&\textbf{4.83}&$6.38$&$5.32$&\textbf{0.92}\\
\hline
Yu et al.~\cite{yu2017learning} &$5.94$&$6.48$&$7.96$&$6.79$&$1.05$\\
\hline
Ours (SIFT)  &$5.65$&$6.23$&$9.24$&$7.04$&$1.93 $\\
\hline
Ours (DFF)  &\textbf{3.68} & $5.03$ &\textbf{5.78} &\textbf{4.83} &$1.06$\\
\hline
\end{tabular}
\end{center}
\end{table}

\begin{figure*}
	\centering
	\includegraphics[totalheight=9.5in]{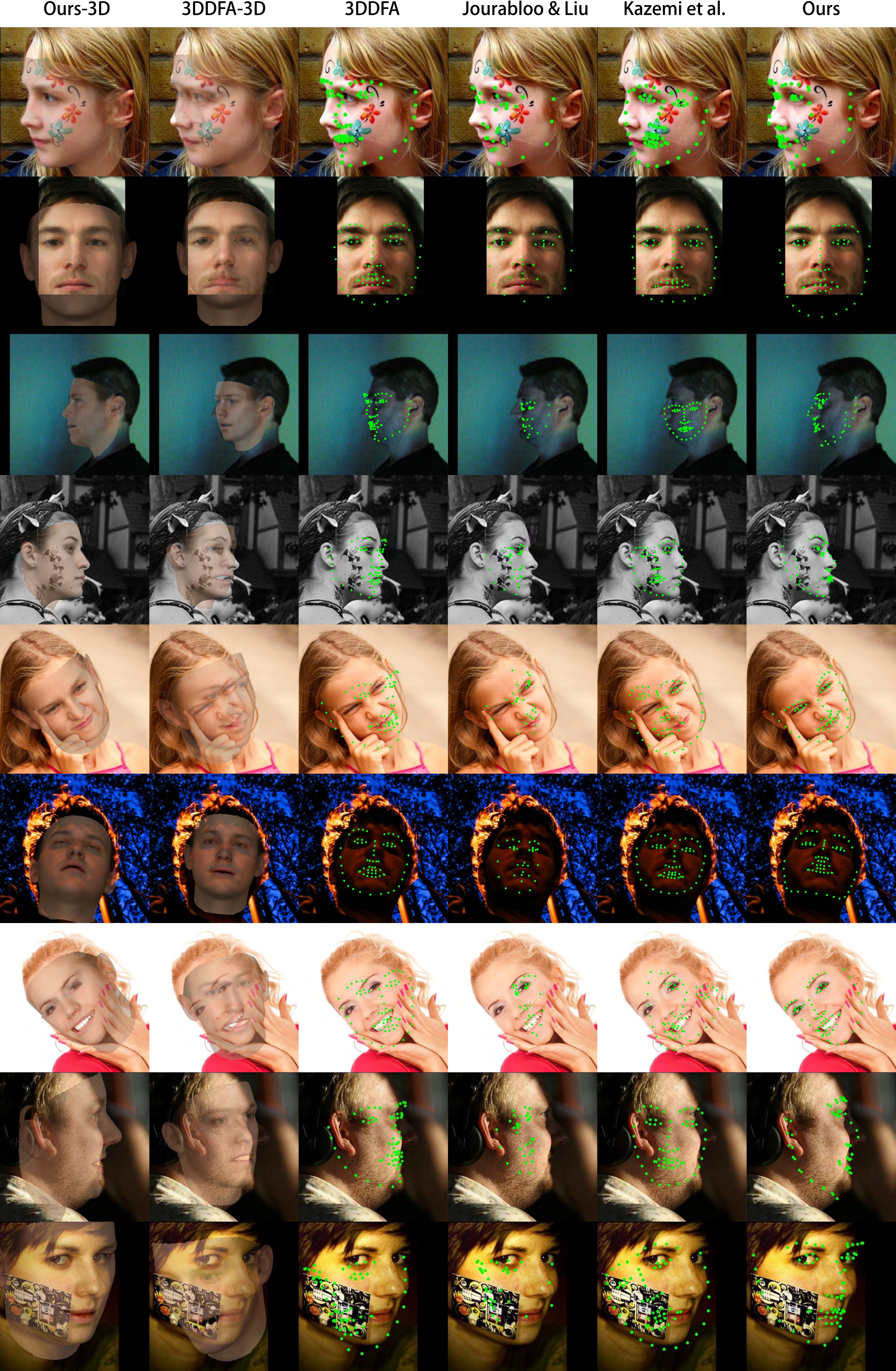}
	\caption{Examples of 2D and 3D facial alignment results on the AFLW2000 dataset~\cite{zhu2016face} using our method, and the methods of Zhu et
		al.~\cite{zhu2016face}, Jourabloo \& Liu~\cite{jourabloo2016large}, and Kazemi et al.~\cite{kazemi2014one}.}
	\label{fig:comparason}
\end{figure*}

\begin{figure*}
	\centering
	\includegraphics[width=\textwidth]{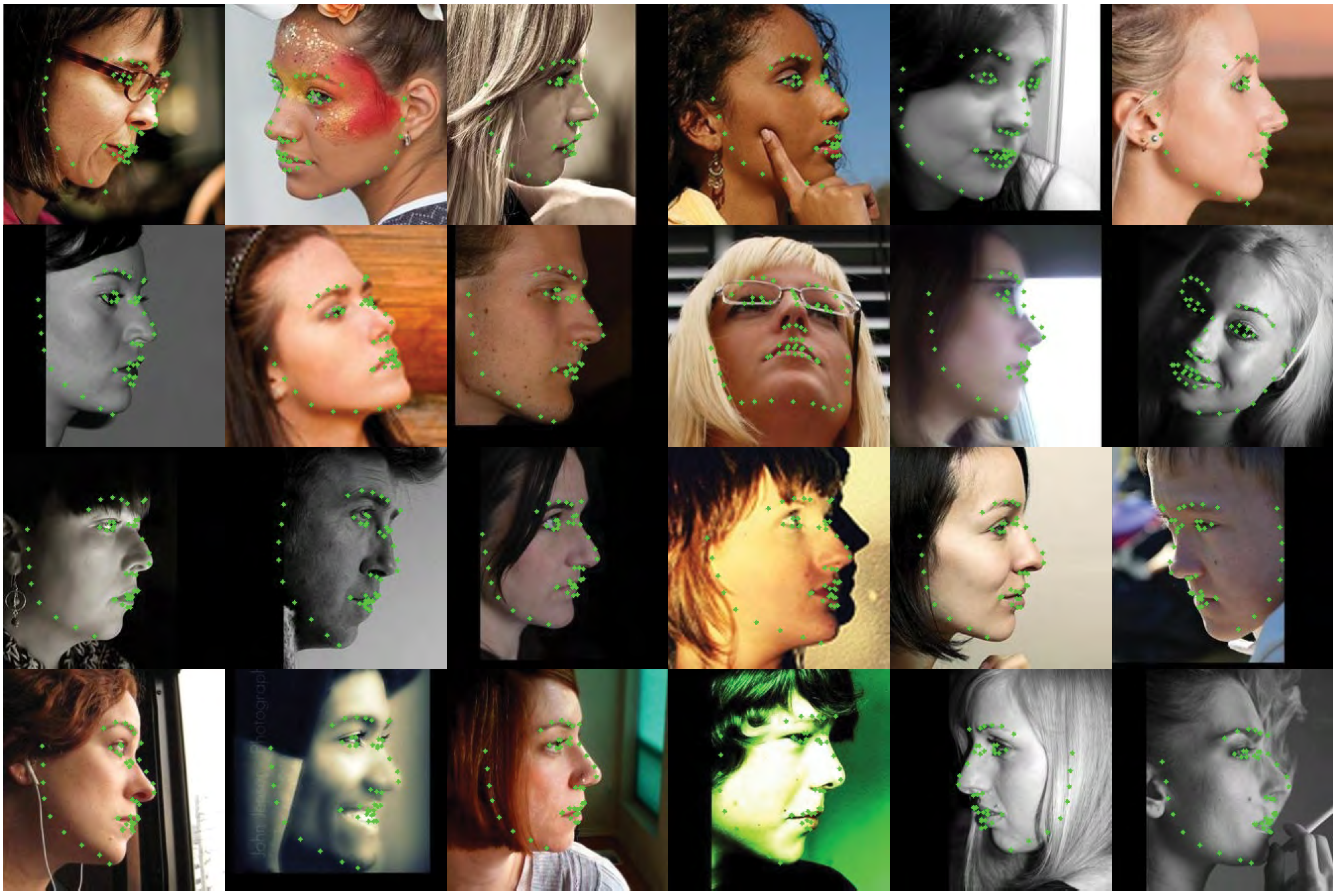}
	\caption{
		Examples of face alignment on large poses from the AFLW database. Only visible landmarks are showed.}
	\label{fig:aflw result}
\end{figure*}

Fig.~\ref{fig:comparason} shows some results of our 2D and 3D facial alignment on challenging and large pose images from AFLW2000. We also show the results using the large pose face alignment methods from~\cite{zhu2016face,jourabloo2016large}, and the widely used state-of-the-art face trackers
of Kazemi et al.~\cite{kazemi2014one}. It can be observed that our method is more robust to heavy occlusions, large variations in illumination, translation, and rotation. More alignment results by our method on AFLW dataset are given in Fig.~\ref{fig:aflw result}.

\begin{table}[!t]
	\caption{The NME ($\%$) on AFLW2000-3D Dataset (68 Points)}
	\label{tab:AFLW2000}
	\begin{center}
		%\scalebox{0.9}{
		\begin{tabular}{|c || c | c | c | c | c |}
			\hline
			Method &[0,30]&[30,60]&[60,90]&Mean&Std\\
			\hline
			RCPR~\cite{burgos2013robust} (300W-LP)  &$4.26$&$5.96$&$13.18$&$7.80$&$4.74$\\
			\hline
			ESR~\cite{cao2014face} (300W-LP)  &$4.60$&$6.7$&$12.67$&$7.99$&$4.19$\\
			\hline
			SDM~\cite{xiong2013supervised} (300W-LP)  &$3.67$&$4.94$&$9.76$&$6.12$&$3.21$\\
			\hline
			3DDFA~\cite{zhu2016face}  &$3.78$&$4.54$&$7.93$&$5.42$&$2.21$\\
			\hline
			3DDFA+SDM  &$3.43$&\textbf{4.24} &$7.17$&$4.94$&$1.97$\\
			\hline
			Yu et al.~\cite{yu2017learning} &$3.62$&$6.06$&$9.56$&$6.41$&$2.99$\\
			\hline
			Ours (SIFT)  &$5.35 $&$6.75 $&$8.23 $&$ 6.78$&\textbf{1.44}\\
			\hline
			Ours (DFF)  &\textbf{3.20} &$4.68$&\textbf{6.28} & \textbf{4.72} & 1.54\\
			\hline
		\end{tabular}
		%}
	\end{center}
\end{table}

\begin{table}[!t]
	\caption{The NME ($\%$) on AFLW2000 for Visible Inner Landmarks}
	\label{tab:AFLW2000_visi}
	\begin{center}
		\begin{tabular}{|c||c|c|c|c|c|}
			\hline
			Method &[0,30]&[30,60]&[60,90]&All Images\\
			\hline
			Zhu et al.~\cite{zhu2016face}  &$4.30$&$4.41$&$6.68$&$4.60$\\
			\hline
			Yu et al.~\cite{yu2017learning}  &$3.14$&$3.84$&$5.53$&$3.58$\\
			\hline
			Ours (DFF)  &\textbf{2.56} &\textbf{3.80}&\textbf{4.80} & \textbf{3.14}\\
			\hline
		\end{tabular}
	\end{center}
\end{table}

\begin{table}[!b]
	\caption{The NME($\%$) of Face Alignment Results on 300W, With the First and the Second Best Results Highlighted}
	\label{tab:Medium_Pose}
	\begin{center}
		\begin{tabular}{|c||c|c|c|c|c|c|}
			\hline
			Method &Common&Challenging&Full\\
			\hline
			TSPM~\cite{zhu2012face}  &$8.22$&$18.33$&$10.20$\\
			\hline
			ESR~\cite{cao2014face}  &$5.28$&$17.00$&$7.58$\\
			\hline
			RCPR~\cite{burgos2013robust}  &$6.18$&$17.26$&$8.35$\\
			\hline
			SDM~\cite{xiong2013supervised}  &$5.57$&$15.40$&$7.50$\\
			\hline
			LBF~\cite{ren2014face}  &$4.95$&$11.98$&$6.32$\\
			\hline
			CFSS~\cite{zhu2015face}  &\textbf{4.73}&\textbf{9.98}&\textbf{5.76}\\
			\hline
			3DDFA~\cite{zhu2016face}  &$6.15$&$10.59$&$7.01$\\
			\hline
			3DDFA+SDM~\cite{zhu2016face}  &$5.53$&\textbf{9.56}&6.31\\
			\hline
			Ours (SDM+DFF)  &\textbf{4.91}&$11.61$&\textbf{6.22}\\
			\hline
		\end{tabular}
	\end{center}
\end{table}

We also test the performance of DFF on medium pose face alignment problems. The experiments are conducted on the 300W dataset, and we use the training data of the LFPW and HELEN datasets, and the whole AFW dataset, to train the alignment model. The testing is performed on three parts: the test samples from LFPW and HELEN as the common subset, the 135-image IBUG as the challenging subset, and the union of them as the full set ($689$ images in total), following the setting of~\cite{zhu2016face}. Since the 300W dataset does not provide ground-truth camera parameters, the large-pose face alignment method described in Sec.~\ref{sec:Large_Method} can not be used. Instead, we assume all landmarks are visible, and apply the SDM method~\cite{xiong2013supervised} to regress the landmark positions in a cascaded way, using DFF instead of SIFT as the feature descriptor.
Each training image is augmented with 5 random face box, resulting in training images 15740 in total. The alignment accuracy is evaluated by the landmark NME, using the inter-pupil distance for normalization. Tab.~\ref{tab:Medium_Pose} compares our results with the results reported in~\cite{zhu2016face}. It can be observed that our method greatly improves the performance of SDM by simply replacing SIFT with DFF, achieving top performance for both the common and full test sets, and close to the top result for challenging test set.

The accompanying video\videoinfo{} shows face feature tracking results using our method, on video clips with speech, poor lighting, large poses, and sports scenes. For comparison, we also show the results using \cite{zhu2016face,jourabloo2016large,kazemi2014one}. Some parts of the video clips with fast face movements are slowed down for more clear comparison. Our results are consistently more robust and accurate for these challenging scenes.

\subsection{Computation Time}
Our experiments are run on a PC with an Intel Core i7-4790 CPU at 4.0GHz, 8GB RAM, and a GTX 1070 GPU.
It takes about 5ms to extract DFF for a $224\times224$ image, and about 8.8ms for each iteration of method described in~\ref{sec:Large_Method}. All the results by our method are generated using three iterations. As we only need to extract DFF once for each image, the total running time is less than 35ms.
Our method takes less computation time than the method in~\cite{zhu2016face}, while achieving more accurate alignment results.

\section{Conclusions}
We present a deep learning based method to extract features from face images. Using a novel feature training method that utilizes the ground-truth correspondence between face images under different poses and expressions in the training set, the resulting deep face feature (DFF) has similar values for the same semantic pixels from different face images.
As a result, the DFF captures global structural information of face images, and is more effective than general feature descriptors like SIFT on face related tasks such as matching and alignment.
We propose a new face alignment method based on the DFF, which achieves state-of-the-art results for large-pose face alignment.
In the future, it would be interesting to explore the use of DFF in other face related tasks.

\bibliographystyle{IEEEtran}
\bibliography{DFF}

\end{document}